\documentclass[letterpaper, 10 pt, conference]{ieeeconf}  

\IEEEoverridecommandlockouts                              

\usepackage{amsmath} 
\usepackage{amssymb}  
\usepackage{hyperref}
\hypersetup{colorlinks = true,
            urlcolor = blue,
            linkcolor = black}
\usepackage{algorithm}
\usepackage[noend]{algpseudocode}
\newtheorem{theorem}{Theorem}[section]
\usepackage{textcomp}
\let\labelindent\relax
\usepackage{enumitem}
\usepackage{multicol}
\usepackage{multirow}
\usepackage{pbox}
\usepackage{caption}
\usepackage{subcaption}
\usepackage{graphicx}
\graphicspath{{./}}
\usepackage{epstopdf}
\title{\LARGE \bf
Planning with Selective Physics-based Simulation for Manipulation Among Movable Objects
}

\author{Muhammad Suhail Saleem$^{1}$ and Maxim Likhachev$^{1}$
\thanks{This work was supported by ONR grant N00014-18-1-2775}
\thanks{$^{1}$Muhammad Suhail Saleem and Maxim Likhachev are with the Robotics Institute, School of Computer Science, Carnegie Mellon University, Pittsburgh, Pennsylvania, USA. {\tt\small \{msaleem2,mlikhach\} @andrew.cmu.edu}}%
}

\begin{document}

\maketitle
\thispagestyle{empty}
\pagestyle{empty}

\begin{abstract}

Use of physics-based simulation as a planning model enables a planner to reason and generate plans that involve non-trivial interactions with the world. For example, grasping a milk container out of a cluttered refrigerator may involve moving a robot manipulator in between other objects, pushing away the ones that are movable and avoiding interactions with certain fragile containers. A physics-based simulator allows a planner to reason about the effects of interactions with these objects and to generate a plan that grasps the milk container successfully. The use of physics-based simulation for planning however is underutilized. One of the reasons for it being that physics-based simulations are typically way too slow for being used within a planning loop that typically requires tens of thousands of actions to be evaluated within a matter of a second or two. In this work, we develop a planning algorithm that tries to address this challenge. In particular, it builds on the observation that only a small number of actions actually need to be simulated using physics, and the remaining set of actions, such as moving an arm around obstacles, can be evaluated using a much simpler internal planning model, e.g., a simple collision-checking model. Motivated by this, we develop an algorithm called Planning with Selective Physics-based Simulation that automatically discovers what should be simulated with physics and what can utilize an internal planning model for pick-and-place tasks.

\end{abstract}

\section{INTRODUCTION}
With the research focus in most domains shifting towards real-world problems, there have been considerable efforts in developing accurate physics-based simulators. Modern-day simulators are capable of modeling complex multi-body interactions faster than real-time. They play a significant role in fields like reinforcement learning where an agent tries to learn a policy by trying out a large number of actions in a simulated environment. However, their impact in the field of planning is still limited.


\begin{figure}
    \centering
    \includegraphics[scale = 0.22]{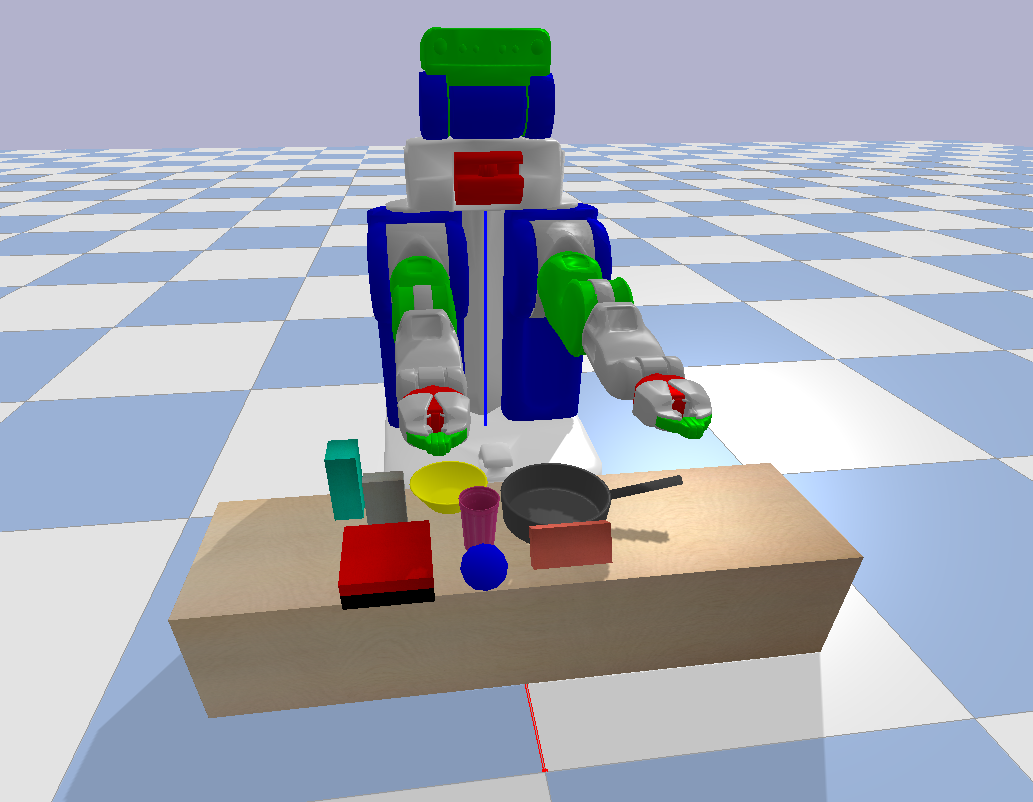}
    \caption{Standard domestic scene where picking up any object would require the robot to interact with other objects.}
    \label{fig:my_label}
\end{figure}
Most planners typically find solutions to planning problems by evaluating the effect of actions from different states using a simple internal planning model like a collision checker. Certain domains like manipulation among movable objects where the robot is allowed to interact with the objects in its environment, require more than a simple planning model. The planner in these cases needs to reason about the dynamics of interactions between the robot and the objects and the object-object interactions they lead to. This necessitates the usage of physics-based simulations. However, their usage by planners is an underexplored area owing to three major challenges. The first being the computational complexity of the simulations. Computing the dynamics of interactions is a complex task that takes substantially longer when compared to simple collision checking. A typical planning problem would involve tens of thousands of action evaluations. Using the simulator to evaluate them all would imply that the planner takes several minutes before coming up with a valid solution. The second challenge is fidelity, i.e. how accurately the simulator can imitate reality. Approximations and assumptions made by the simulators to speed up computations have an effect on its fidelity. The third being the assumption that the physics-based simulator is aware of the accurate model and parameters (eg. masses) of all objects that are being simulated.

In this work we address the first challenge for the domain of manipulation planning among movable objects through non-prehensile maneuvers. While the simulation itself is slow, not all actions have to be evaluated through simulation. We use a simple internal collision checking model to evaluate the majority of actions while intelligently choosing which actions to simulate without affecting completeness guarantees. We apply our approach to the 7 DOF manipulator on Willow Garage's PR2 robot for several complex and cluttered environments. Simulation results have shown that our technique is capable of producing bounded suboptimal paths rather quickly despite using simulations. The utility of the algorithm was also demonstrated through real world experiments on the 7 DOF manipulator of UBTech's Walker robot.

\section{RELATED WORK}
Navigation among Movable Obstacles (NAMO) is the class of problems where the robot is allowed to move objects in its environment to perform its intended higher level task. This is especially useful in manipulation planning as in a cluttered environment the free space is extremely constrained and in most cases there is an absence of a collision free path from start to goal. However, NAMO is an NP-hard problem which can not be tackled through brute force \cite{Wilfong} \cite{PushPush}.

Modelling and solving manipulation planning in clutter as a motion planning problem is a difficult task owing to the high dimensional search space. As the objects are movable, their degrees of freedom become part of the search state thereby making the problem intractable through traditional planning techniques. Previous approaches made assumptions like predefining contact points and the final configurations of all movable objects \cite{ota2004rearrangement}\cite{alami1994two}\cite{stilman2005navigation} to help simplify the problem. Stilman et al. assumed monotone plans \cite{stilman2007manipulation}, which implies that if there exists a solution, it could be found by moving every obstacle at most once. While these assumptions ease the planning problem, they sacrifice completeness guarantees.

As discussed by Dogar et al. in \cite{dogar2012physics}, most of the initial approaches which aimed at rearranging clutter took an object centric approach \cite{dogar2011framework}\cite{stilman2007manipulation}. They tried to move one object at a time while ensuring no contact was established with the other objects. Further, they restricted their interactions with the objects to purely prehensile maneuvers. These approaches proved highly inefficient and slow.

One of the first steps towards the use of non-prehensile maneuvers was taken by Mason and Lynch, who studied the mechanics of pushing. They developed a planner which could stably push objects \cite{mason2001mechanics}\cite{lynch1996stable}. Goyal et al predicted the the effect of pushing through \textit{limit surfaces} in \cite{goyal1991planar}. More recent works tried to generalize the problem of rearranging clutter to non-prehensile actions \cite{dogar2011framework}\cite{dogar2012planning}\cite{king2016robust}. Push grasping maneuvers were introduced in \cite{dogar2010push}, which were later applied to cluttered environments to create collision free paths for the manipulator \cite{dogar2011framework}.

The usage of physics-based simulations in planning is heavily underutilized. The approach introduced in \cite{dogar2012physics}, uses physics-based simulations to precompute and cache robot-object interactions. However, they restrict object-object interactions to prevent online simulations and make the problem computationally tractable. \cite{moll2017randomized} and \cite{king2016robust} interleave a sampling-based kinodynamic motion planner with a physics-based simulator in an online fashion. They attempt to tackle the challenge of low fidelity in simulations, by accounting for the pose uncertainties of the interacted objects. The focus of our work is tangential. We aim to speed up planning when integrated with a slow but high fidelity simulator.

\section{PROBLEM FORMULATION}

Our domain consists of a robot manipulator $\mathcal{R}$ of $d \in \mathbb{N}$ degrees of freedom and an environment consisting of $n$ objects ($n \in \mathbb{N}$) represented by $\mathcal{O} = \{O_1,O_2...O_n\}$. The state space of the robot is represented by $X_R \subset \mathbb{R}^d$. The set of all actions it can execute, is defined by its action space $\mathcal{A}_R$. Similarly, every object $O_k \in \mathcal{O}$ has a state space $X_{O_k} \subset \mathbb{R}^3 \times SO_3$, defined by their position ($x,y,z$) and orientation ($roll,pitch,yaw$). Therefore, we can define the state space of the planner $\mathcal{S}$, as the Cartesian product of the spaces of the manipulator and the objects, $\mathcal{S} = X_R \times X_{O_1} \times X_{O_2} ... \times X_{O_n}$. A state $s \in \mathcal{S}$ is given by  $s = \{x_R, x_{O_1}, x_{O_2} .. x_{O_n}\}$ where, $x_R \in X_R$ and $x_{O_k} \in X_{O_k}$.

 \subsection{Interaction Constraints}
We define \textit{interaction} as a collision between two rigid bodies (object-object or robot-object) that arises as a result of the robot\textquotesingle s actions. It has to be noted that interactions are defined between pairs of bodies. In a case where three objects are colliding with each other simultaneously, we account for it as three different interactions.

The robot is allowed to interact with the objects in its environment subject to certain conditions. These conditions are referred to as \textit{interaction constraints}. The constraints are defined with respect to the objects in the environment and when they are violated, they are said to be violated with respect to the object. These constraints can be used to determine the validity of an action $a \in \mathcal{A}_R$ taken from a state $s \in \mathcal{S}$. They could be formulated in different ways based on the problem we would like to address. Some examples include:
 \begin{itemize}
     \item Disallowing interactions with certain objects $O_k \in \mathcal{O}$.
     \item Disallowing interactions that cause certain objects $O_k \in \mathcal{O}$ to topple and fall over. A cup of coffee or a glass of water are examples of objects that we would not like to topple.
 \end{itemize}
We provide a generic framework that is compatible with a wide array of constraints.

 \subsection{Problem Statement}

 \textit{To determine a path $\pi$ defined as a sequence of robot states $x_R \in X_R$, that takes the robot from the initial state to a goal specified in the end-effector workspace $X_G$ (usually a pregrasp location), while minimizing a cost function (path length) and respecting the interaction constraints.}

 Unlike previous approaches, our definition of goal places no constraints on the final configuration of the objects (except the target object, which is assumed to be fixed as the pregrasp location is static). This implies $X_G$ represents a goal region, $X_G \subset \mathcal{S}$. This also increases the complexity of the problem as the search now has to reason about configurations of objects that do not hinder the robot from reaching any $x_{G} \in X_{G}$.

 \subsection{Assumptions}
 We classify interactions into two types, \textit{first order} and \textit{second order}. First order interactions are defined as direct interactions between the robot $\mathcal{R}$ and an object $O_k \in \mathcal{O}$. Second order interactions are defined as interactions between two bodies that arises as a result of a first order interaction. For example, an object ${O}_i$ colliding with another object ${O}_j$ as a result of a robot action toppling ${O}_i$ (as seen in Fig. 2). Given these definitions, we make the following assumption:
 \begin{enumerate}[label=\textbf{{A}{{\arabic*}})}]
 \hypertarget{Assumption1}{\item} \textit{If a problem has valid solutions, there exists at least one solution that does not cause any second order interactions.}  \end{enumerate}
 This is a reasonable assumption as in most practical cases the robot can always find a sequence of actions that can move every object individually, while ensuring no contact with the rest of the objects. This assumption is purely from the standpoint of theoretical properties. While our algorithm is capable of finding solutions that cause both first order and second order interactions, the proof of completeness as explained in section V holds under this assumption.
\begin{figure}[htbp]
  \centering
  \includegraphics[scale = 0.65]{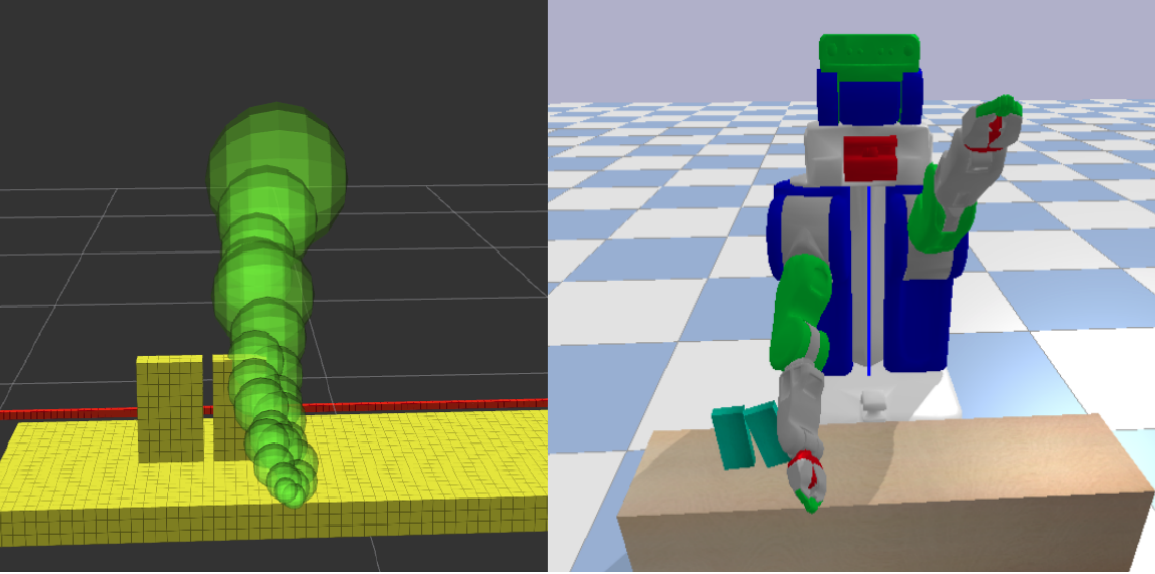}
  \caption{Simple internal collision checking model (left) vs external physics-based simulator (right).}
  \label{checkervssimulator}
\end{figure}
\vspace{-0.5cm}
\subsection{Forward Simulations}
Given interaction constraints, planners reason about actions based on a model of the world. If the model is physics-based and accounts for the dynamics of interactions and their effects, action evaluations are referred to as \textit{external simulations} due to their requirement of an external physics engine. However, if the model is a simple internal model that can purely detect collisions, action evaluations are referred to as \textit{collision checks}. Collision checks can only detect interactions between rigid bodies and not evaluate their dynamics or their effects.

For example, when a manipulator interacts with an object in a clutter, a first order interaction between the robot and the object could cause it to fall over another object as shown in Figure \ref{checkervssimulator}. The collision checker can only detect the first order interaction. However, simulations can detect both the first order interactions and the second order object-object interactions it causes.

Evaluating an action $a \in \mathcal{A}_R$ from a state $s \in \mathcal{S}$ using either of these models, will give us the following information:
        \begin{itemize}
            \item $P(s,a) = \{(x,y)$ : $x,y \in \mathcal{O} \cup \{\mathcal{R}\} $, pair of bodies interacting with each other when evaluating action $a$ from state $s$ using a specific model\}
            \item $O(s,a) \subseteq \mathcal{O}$ = Set of all unique objects in $P(s,a)$
            \item State $s' \in \mathcal{S}$ reached by executing the action
        \end{itemize}
It is to be noted that a single action can result in multiple interactions making $P(s,a)$ a set of pairs of bodies.
If we use an external physics engine, the notations are $P_E(s,a)$, $O_E(s,a)$ and $s'_{e}(s,a)$ and if the simulation is through a collision checker, the notations are $P_C(s,a)$, $O_C(s,a)$ and $s'_{c}(s,a)$. Since all interactions detected through collision checkers are detected through external simulations as well,
$ P_C(s,a) \subseteq P_E(s,a)$ and $ O_C(s,a) \subseteq O_E(s,a)$.

\section{ALGORITHM}
Our approach can be divided into two major components. The first being detailed in
\hyperlink{alg1}{Algorithm 1}, is a framework for intelligently adapting any search based planner for the task of manipulation in clutter through non-prehensile manipulation. However, using \hyperlink{alg1}{Algorithm 1} naively with search based planners would be highly inefficient. The second component as discussed in \hyperlink{alg2}{Algorithm 2}, is an approach that allows you to selectively simulate actions without losing completeness and suboptimality guarantees. This, when integrated with \hyperlink{alg1}{Algorithm 1}, speeds up planning multifold.

\hypertarget{alg1}{\begin{algorithm}[H]}
\caption{Get Successors Routine}
\label{alg:amha}
\begin{algorithmic}[1]
\Procedure{GetSuccs}{s,a}
\State {$P_C(s,a), O_C(s,a), s'_{c}$ = \textit{collision check} ($s,a$)}
\State \textbf{If }{$O_C(s,a)$ = $\emptyset$}
\State $\>$ Return $s'_{c}$ \Comment{valid successor}
\State \textbf{Else if }{constraint violation \Comment{\textit{collision check}}}
\State $\>$ Return invalid
\State \textbf{Else if } {$\exists O_k \in O_C(s,a)$ $\mid$ $O_k \in C_{R}$}
\State $\>$ $P_E(s,a), O_E(s,a), s_{e}$ = \textit{ext. simulation} ($s,a$)
\State $\>$ \textbf{If } {constraint violation \Comment{\textit{ext. simulation check}}}
\State $\>$ $\>$ Return invalid
\State $\>$ \textbf{Else}
\State $\>$ $\>$ Return $s'_{e}$ \Comment{valid successor}
\EndProcedure
\end{algorithmic}
\end{algorithm}


\hypertarget{alg2}{\begin{algorithm}}
\caption{Planning with Selective Simulation}
\label{alg:amha_2}
\begin{algorithmic}[1]
\Procedure{Track}{path}
\State $P = \emptyset$
\State $s = s_0$ \Comment{$s_0 \in \mathcal{S}$ is the start state}
\For{every action $a \in$ path}
\State $P_E(s,a), O_E(s,a), s'_{e}$ = \textit{ext. simulation} ($s,a$)
\State $P = P \cup P_E(s,a)$
\State \textbf{If }{Interaction constraint was violated at $O_v \in \mathcal{O}$}
\State $\>$ $C_{R} = C_{R} \cup $ \Call{RelevantObject}{$P, O_v$}
\State $\>$ Return failure
\State $s$ = $s'_{e}$
\EndFor
\State Return success
\EndProcedure
\Procedure{RelevantObject}{$P, O_v$}
\State $s_{start}$ = $O_v$
\State $Queue = \emptyset$
\State Insert $s_{start}$ into $Queue$
\State \textbf{loop} \Comment{Tree search}
\State $\>$ remove $s$ the front element in $Queue$
\State $\>$ \textbf{If} ($s \notin C_R$)
\State $\>$ $\>$ Return s
\State $\>$ explored[s] = true
\State $\>$ \textbf{for} every $(s,O_k) \in P$ where $O_k \in \mathcal{O}$ \textbf{do}
\State $\>$ $\>$ \textbf{If} (not explored[$O_k$])
\State $\>$ $\>$ $\>$ Insert $O_k$ into $Queue$
\EndProcedure
\Procedure{Main}{\null}
\State $C_{R} = \emptyset$; \Comment{$C_R$ declared globally}
\State \textbf{loop}
\State $\>$ Add all $O_k \in C_{R}$ to \textit{ext. simulator}
\State $\>$ Add all $O_k \in C_{R}$ to \textit{collision checker}
\hypertarget{plan}{\State $\>$ path = \Call{Plan}{\null}}
\State $\>$ \textbf{If} path exists
\State $\>$ $\>$ \textbf{If} $C_{R} \neq \mathcal{O}$
\State $\>$ $\>$ $\>$ Add all $O_k \in \mathcal{O}$ to \textit{ext. simulator}
\State $\>$ $\>$ $\>$ \textbf{If} \Call{Track}{path} is success
\State $\>$ $\>$ $\>$ $\>$ Return path
\State $\>$ $\>$ \textbf{Else}
\State $\>$ $\>$ $\>$ Return path \Comment{Returned path will be valid as}
\State $\>$ \textbf{Else} $\>$$\>$$\>$ $\>$$\>$$\>$$\>$$\>$$\>$$\>$$\>$$\>$$\>$$\>$$\>$$\>$$\>$$\>$$\>$$\>$$\>$$\>$$\>$$\>$$\>$$\>$$\>$$\>$$\>$$\>$  all objects were considered
\State $\>$ $\>$ Return no solution exists
\EndProcedure

\end{algorithmic}
\end{algorithm}

\subsection{Get Successors}
\textit{GetSuccessors} is a standard routine used in search based planners. This routine determines the successor states $s'$ that can be reached from a state $s \in \mathcal{S}$ given the action space $\mathcal{A}_R$ and the interaction constraints. In most domains, the goal is to construct collision free paths. Hence, the interaction constraint is to not interact with any object $O_k \in \mathcal{O}$. In these cases there are no second order interactions as the first order interactions which cause them are disallowed. Hence, collision checks suffice.

However, our domain allows robot-object and object-object interactions. Collision checkers by themselves can not be used to evaluate the validity of all actions as they can not evaluate second order interactions. While external simulations can, they are substantially slower. We combine the two by using collision checkers to detect first order interactions and call the external simulator only if the collision checker has detected an interaction that does not violate the constraints.

For example, let the interaction constraints be to not cause any first-order or second-order interactions with certain fixed objects $\mathcal{F} \subseteq \mathcal{O}$. Any action that leads to first-order interactions with fixed objects $O_k \in \mathcal{F}$ is invalid and any action that does not lead to any first-order interactions is trivially valid.  The external simulator need not be queried in these cases. The external simulator can be queried only when the collision checker detects a first-order interaction with movable objects $O_k \notin \mathcal{F}$. The simulator is then used to determine the validity of the action by evaluating the second order interactions the action causes.

\hyperlink{alg1}{Algorithm 1} presents a generic routine applicable to a range of constraints. Information about the interaction constraints could be leveraged to further optimise the calls to the simulator. $C_R$ in \hyperlink{alg1}{Algorithm 1} represents the set of relevant objects considered by the planner. If we are to plan in the presence of all objects, $C_R = \mathcal{O}$. In the following subsections we formally introduce the concept of relevant objects and discuss an effective method to compute $C_R$, to reduce the number of simulation queries.

\subsection{Relevant Object Based Selective Simulation}

Since a clutter contains a large number of objects, a significantly high number of action evaluations lead to valid first-order interactions and have to be externally simulated. Further, as defined previously, the state space of the problem is given by $\mathcal{S} = X_R \times X_{O_1}... \times X_{O_n}$.  This implies that the search space is extremely large for a cluttered scene. These problems introduce the requirement for intelligent optimizations.


While there are several objects in a cluttered scene, only a few impact our plan. Planning by only considering these objects is sufficient to generate a valid path.
We refer to these objects as \textit{relevant objects}, while the rest of the objects are referred to as \textit{irrelevant objects}. Considering only relevant objects while planning implies that instead of simulating actions that result in first-order interactions with any object, we selectively simulate actions that result in interactions only with the relevant objects. This significantly reduces the number of simulator queries as in most practical cases the number of relevant objects is much lesser than the total number of objects in the scene. However, the identification of these objects is a non-trivial task.


The idea of doing a search in a low dimensional space except in places where the low dimensional search fails was introduced by Gochev et al. \cite{gochev2011path}. This technique speeds up search considerably, especially for high dimensions. We extend this idea to our domain by identifying the relevant objects in a recursive manner as outlined in \hyperlink{alg2}{Algorithm 2}.

Let $\mathcal{S}_{sel}$ represent the search space for our approach and $\pi_{sel}$ be the plan generated by a search in this space. $C_R$ is the set of identified relevant objects which are included in the planning process, $C_R = \{ c_{R_1}, c_{R_2}..c_{R_m} \}$, $c_{R_k} \in \mathcal{O}$, $m \leq n$ .
This implies the set of irrelevant objects $C_I$, is given by $C_I = \mathcal{O} \backslash C_R$. $\mathcal{S}_{sel}$ is a projection of $\mathcal{S}$ onto a lower dimensional space. A state $s_{sel} \in \mathcal{S}_{sel}$ is given by $s_{sel} = \{x_R, x_{c_{R_1}},x_{c_{R_2}}..x_{c_{R_m}}\}$. The low dimensional state $s_{sel}$, maps to a set of states in the high dimensional space $\mathcal{S}$, corresponding to all possible configurations of all irrelevant objects $O_k \in C_I$ for a specific configuration of the robot and the relevant objects $O_k \in C_R$. This many-to-one mapping from $\mathcal{S}$ to $\mathcal{S}_{sel}$ is represented by $\lambda$.

\vspace{-0.4cm}
\begin{equation}
    \lambda: \mathcal{S} \rightarrow \mathcal{S}_{sel}
\vspace{-0.2cm}\end{equation}

We initially assume that no object in the environment is relevant, ie. $C_R = \emptyset$ and $C_I = \mathcal{O}$. $\mathcal{S}_{sel}$ for this planning iteration is equivalent to the state space of the robot $\mathcal{X}_R$.
As the search space is extremely limited and the external simulator is never queried as $C_R$ = $\emptyset$ (\hyperlink{alg1}{Algorithm 1}), $\pi_{sel}$ will be generated substantially quickly.
Since none of the objects were included in the planning process, $\pi_{sel}$ feasible in $\mathcal{S}_{sel}$ does not imply it is feasible in $\mathcal{S}$. Hence, to test the validity of $\pi_{sel}$ in $\mathcal{S}$, we add all objects $O_k \in \mathcal{O}$ to the external simulator (line 32) and track the path (lines 1-11). If no interaction constraints were violated when tracking the path, it implies $\pi_{sel}$ is valid in $\mathcal{S}$. However, if the constraints were violated, $\pi_{sel}$ is invalid in $\mathcal{S}$.

\begin{figure*}[htbp]
  \centering
  \includegraphics[width = \textwidth]{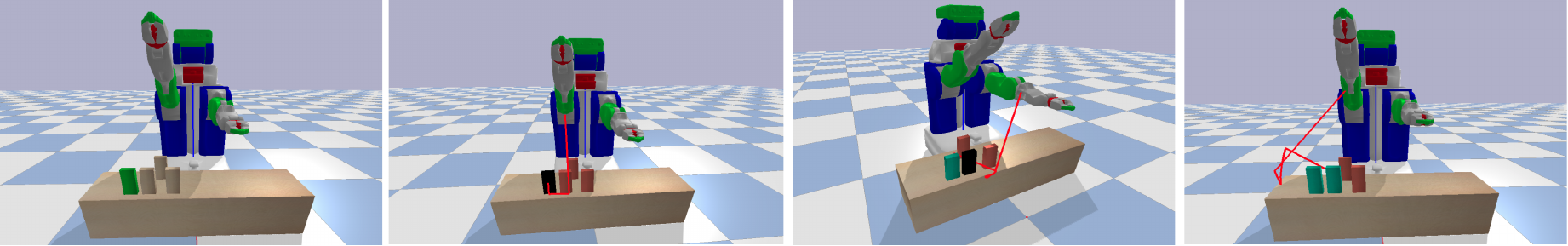}
  \caption{The first image presents a planning scene where the object highlighted in green is forbidden from being interacted with. The next three images represent three planning iterations. Color coding of the objects are as follows.  Red: Irrelevant, Blue: Relevant, Black: Irrelevant but identified as relevant for the next planning iteration. The plan generated in each of these iterations has been highlighted in red.}
\end{figure*}

A plan $\pi_{sel}$ produced by a search in $\mathcal{S}_{sel}$ will ensure the following. In the absence of irrelevant objects $O_k \in C_I$,
\begin{itemize}
    \item No first order interaction between the robot $\mathcal{R}$ and the identified relevant objects $C_R$, violates the constraints.
    \item No second order interactions between the identified relevant objects $C_R$,  violates the constraints.
\end{itemize}

Thus, for $\pi_{sel}$ to be invalid in $\mathcal{S}$, there should have been interactions with the irrelevant objects $O_k \in C_I$ that were not accounted for during planning. This implies there exists at least one object $O_k \in C_I$ which is relevant and yet to be added to $C_R$.
This object is then identified and added to $C_R$ in the next iteration of planning. Now $C_R = {O_k}$, $C_I = \mathcal{O} \backslash O_k$, and $\mathcal{S}_{sel} = X_R \times X_{O_k}$. This process is continued in a recursive fashion until the returned path $\pi_{sel}$ is valid in $\mathcal{S}$.

As and when we identify more relevant objects we adaptively increase the set of objects we reason about. The action space $\mathcal{A}_R$ is defined with respect to the robot and is independent of the search space. This means that $\mathcal{A}_R$ remains unaltered as we increase the search space every planning iteration. Further, it also means that the path $\pi_{sel}$ requires no additional processing to be executed in $\mathcal{S}$.



\subsection{Relevant Object Identification}

For example, let a scene consist of $n$ objects $\{O_1, O_2,...O_n\}$. The interaction constraint for the problem is to not cause any interactions with $O_1$. For a particular planning iteration let $C_R = \{O_1,O_2\}$. While tracking $\pi_{sel}$ generated in this iteration, let a robot action $a \in \mathcal{A}_R$ cause an irrelevant object $O_3$ to fall over $O_2$ which in turn interacts with $O_1$. Although $O_2$ was the object that interacted with $O_1$, the irrelevant object $O_3$ caused this interaction. Including $O_3$ in the planning process would have prevented the violation.

The goal of this routine is to identify the irrelevant object $O_k \notin C_R$ that caused the constraint violation while tracking $\pi_{sel}$.
For this purpose we construct a tree $T_{obj}$ and do a Breadth First search over it (lines 16-23).
Let $O_v \in \mathcal{O}$ represent the object with respect to which the constraint was violated. The tree $T_{obj}$ is defined by its:
\begin{itemize}
    \item Nodes: $O_k \in \mathcal{O}$
    \item Edges: Interactions that occurred while tracking $\pi_{sel}$
    \item Root Node: $O_v$
\end{itemize}
This means the nodes at depth $d=1$ represent objects that had directly interacted with $O_v$. If no object $O_k \notin C_R$ is at depth $d=1$, we search the second level to identify if any of the relevant objects in the first level had interacted with an irrelevant object. Thus, we continue the search until we identify the first irrelevant object $O_k \notin C_R$ (line 18). Although, the objects at depth $d=1$ were the ones that had directly interacted with $O_v$, as explained above, $O_k$ was the cause of it. $O_k$ is identified as relevant and added to the next iteration of planning (line 8).

\subsection{Planning Algorithm}
Since our algorithm can be integrated with most search based planners, we have not outlined any specific planning routine in \hyperlink{alg2}{Algorithm 2}. Instead, we have a generic call to a planner in \hyperlink{plan}{line 29}, where \hyperlink{alg1}{Algorithm 1} would be used instead of the regular Get Successors routine.

While other algorithms can be used, we have empirically found that
our adaptation of the Lazy Weighted A* (LWA*) search introduced in \cite{Cohen2014} maximizes performance. Details of the search have been mentioned below while the evaluation of the performance can be found in the next section.

Lazy Weighted A* is a variant of Weighted A* (WA*) that was developed for problems where the evaluation of the true cost of some actions is far more expensive than the others. They fully evaluate the cost of these actions only when the planner intends to use them. An underestimate of the original cost of the action is used to add the successor to $OPEN$. Only when the successor is being expanded by the search do they compute the true cost of reaching it. Based on this true cost the successor is reinserted into the search.

We use the same logic to limit the number of calls to the external simulator. For our case the cost of the edge is known, instead the validity of the edge has to be determined. Only when the state is being expanded do we call the external simulator to determine the validity of the action leading to it. If it is valid, we expand the state. If not, we remove the state from the search and expand the next one.

\section{THEORETICAL PROPERTIES}
\hypertarget{theorem1}{\begin{theorem}}
\textit{An action $a \in \mathcal{A}_R$ that is valid from a high dimensional state $s \in \mathcal{S}$ and does not result in second order interactions, is always valid from $\lambda(s) \in \mathcal{S}_{sel}$.}
\end{theorem}
Let $a \in \mathcal{A}_R$ be an action that does not result in second order interactions when executed from a high dimensional state $s \in \mathcal{S}$. Let the set of interactions that arise as a result of the action from $s$, be $P_E(s,a)$ and the set of interactions that arise from its lower dimensional projection $\lambda(s)$, be $P_E(\lambda(s),a)$. Then we can write, $P_E(\lambda(s),a) \subseteq P_E(s,a)$. This is because the objects we consider in $\mathcal{S}_{sel}$ is a subset of the objects considered in $\mathcal{S}$. Thus, if such an action is valid in the higher dimensional state it will always be valid in its lower dimensional projection.


\begin{theorem}
\textit{Assuming \hyperlink{Assumption1}{\textbf{A1}}, if the search in $\mathcal{S}_{sel}$ returns no valid solution, there exists no valid solution in $\mathcal{S}$}
\end{theorem}

From \hyperlink{Assumption1}{\textbf{A1}}, we know there always exists at least one solution that does not result in second order interactions. A solution in $\mathcal{S}$ will be present in $\mathcal{S}_{sel}$ if the actions resulting in the path are not invalidated in $\mathcal{S}_{sel}$. According to Theorem 5.1, all actions that are valid in $\mathcal{S}$ and do not result in second order interactions are valid in $\mathcal{S}_{sel}$. The search in $\mathcal{S}_{sel}$ is guaranteed to find the path that does not lead to second order interactions as such actions are never invalidated in $\mathcal{S}_{sel}$. Hence, if the search returns no solutions it implies no valid solutions exist for the problem. Thus, the algorithm is complete under assumption \hyperlink{Assumption1}{\textbf{A1}}.


\section{EXPERIMENTAL ANALYSIS}

We evaluated the performance of our algorithm in simulation on a 7 DOF PR2 manipulator. Bullet Physics Engine \cite{coumans2019} was used for external simulations and a custom built collision checker was used for the simpler internal evaluations. We evaluated the performance of our Selective Physics-based Simulation algorithm with Weighted A* search \cite{pohl1970heuristic} and Lazy Weighted A* search \cite{Cohen2014}. To demonstrate the effect of selective simulation, this was benchmarked against the same planning algorithms searching without selective simulations.

The experiments were conducted for cluttered tabletop scenes consisting of several household objects as shown in Figure 1. The robot's goal was to reach an ($x,y,z,r,p,y$) pose specified in its end-effector workspace. The interaction constraint was to not cause any first order or second order interactions with specific fixed objects in the scene. The heuristic used was obtained by a 3-D Breadth-First search initialized at the $(x,y,z)$ coordinates of the goal considering only the immovable objects in the environment. This will be an underestimate of the true cost of transition, making the heuristic admissible. The maximum time available for the planners was two minutes. If a planner could not find a solution within this time, we recorded it as a failure. The results presented in Table 1 and Figure 5 are average statistics for 60 experiments. The configurations of all the objects in the scene and the goal pose were randomly generated for each experiment.
\begin{table}
\caption{Performance comparison between the Selective Simulation and Non-Selective Simulation approach. Number of objects in the scene = 12, number of fixed objects = 2.}
\begin{center}
\begin{tabular}{ |p{2.2cm}|p{1.3cm}|p{1cm}|p{1.3cm}|c| }
\hline
 & \multicolumn{2}{|c|}{Selective Simulation} & \multicolumn{2}{|c|}{Non-Selective}\\
 & \multicolumn{2}{|c|}{Search} & \multicolumn{2}{|c|}{Search} \\ \cline{2-5}
Metric & {Lazy WA*} & WA* & Lazy WA* & WA* \\
 \hline
Success Rate & 0.92 & 0.88 & 0.40 & 0.35 \\
\hline
Planning Time (s) & 3.23 & 5.38 & 62.342 & 98.26 \\
\hline
\end{tabular}
\end{center}
\end{table}

 \begin{figure}
\vspace{-15pt}

    \centering
    \begin{subfigure}[b]{0.35\textwidth}
     \centering
     \includegraphics[width=\textwidth]{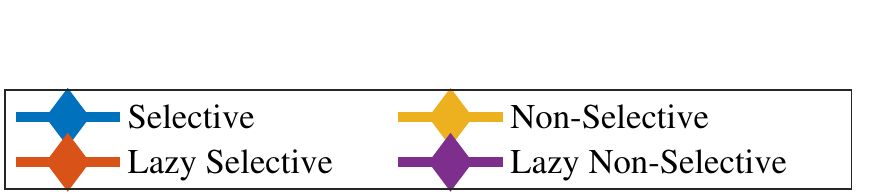}
     \label{fig:run_time_1_b}
    \end{subfigure}\vspace{-5pt}

    \begin{subfigure}[b]{0.23\textwidth}
         \centering
         \includegraphics[width=\textwidth]{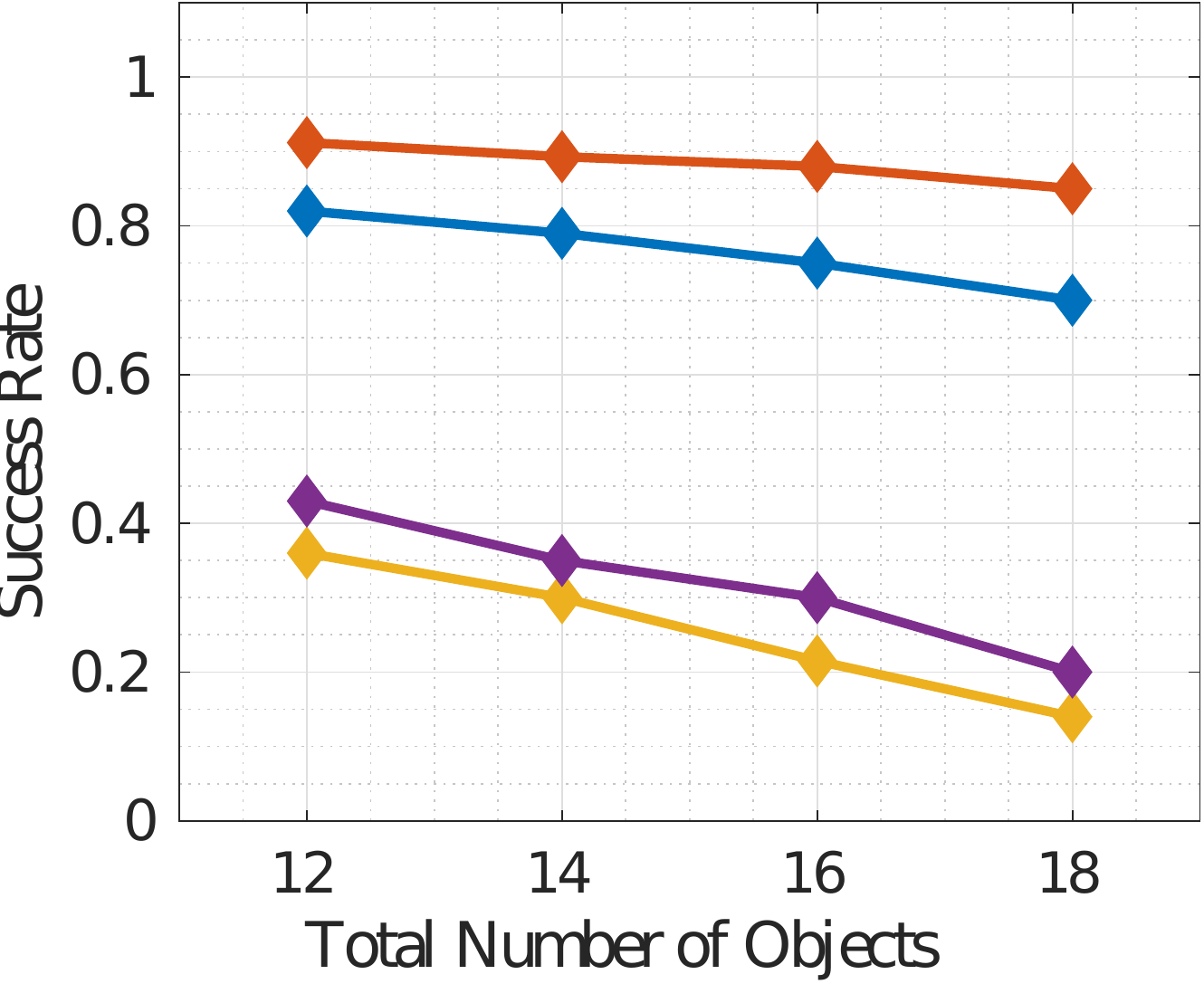}
         \caption{}
         \label{fig:run_time_1_a}
    \end{subfigure}
    \vspace{-0pt}
    \begin{subfigure}[b]{0.23\textwidth}
         \centering
         \includegraphics[width=\textwidth]{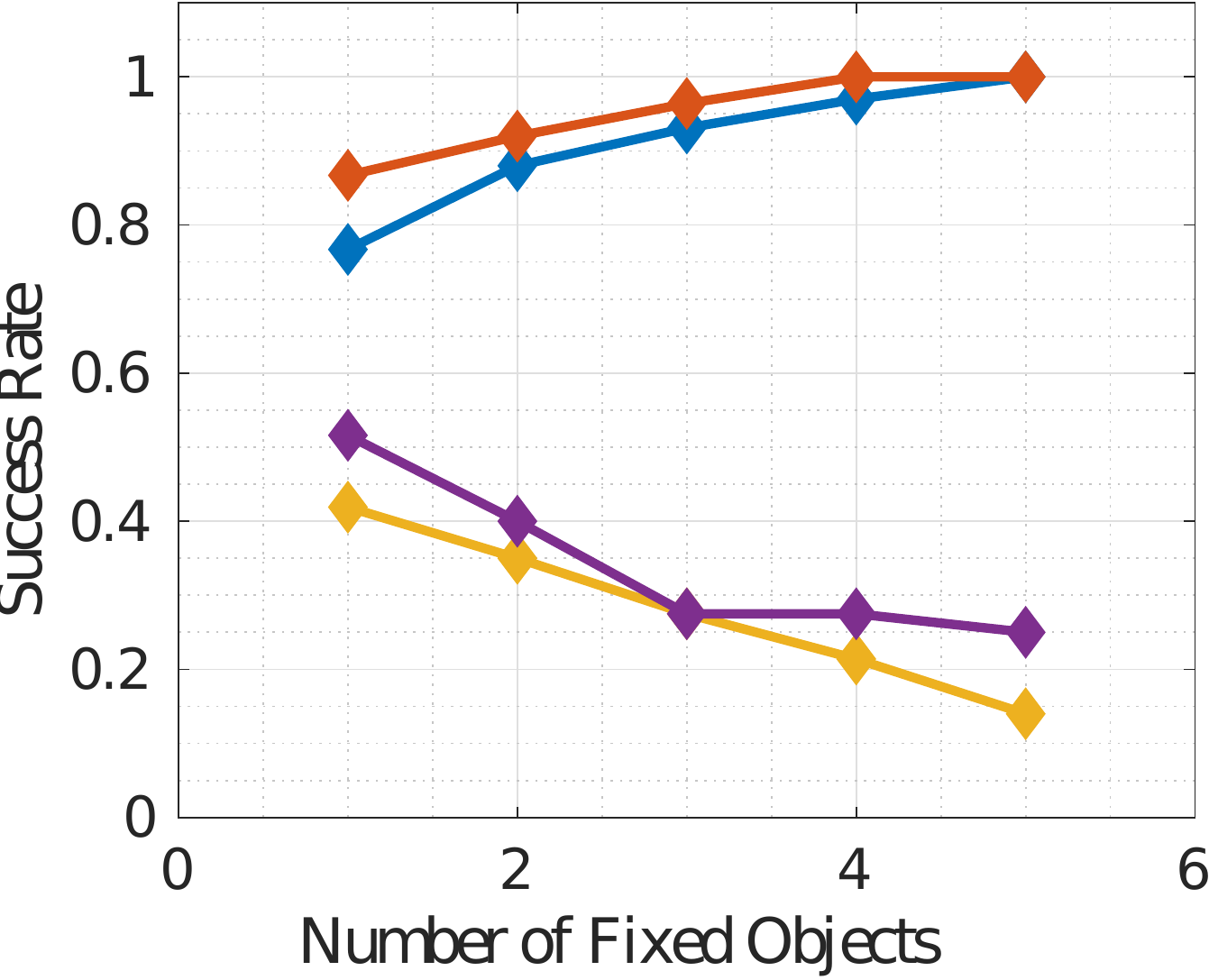}
         \caption{}
         \label{fig:run_time_1_b}
    \end{subfigure}

    \label{fig:run_time_1}

    \caption{(a) Success rate vs total number of objects. Number of fixed objects in the scene = 3 (b) Success rate vs number of fixed objects. Total number of objects in the scene = 12.}
    \vspace{-5mm}
\end{figure}



From Table 1 it is clear that our selective simulation technique in combination with our adaptation of Lazy Weighted A* considerably outperforms the Non-Selective search. We have a much higher success rate with significantly faster planning times. Selective simulation has sped up planning by more than 20 times. This makes the usage of physics based models with planners feasible, opening up a range of possibilities with respect to the domains that can be explored.

Further, as shown in Figure 4a our algorithm's performance remains almost constant as the total number of objects in the scene increases. This is because the number of relevant objects does not scale with the total number of objects in the scene. However, the performance of non-selective search goes down significantly as higher number of actions have to be simulated with an increase in the number of objects. This is an important property that again demonstrates the utility of our algorithm in the real world, as in most domestic cases the total number of objects in the scene are on the higher side. Figure 4b presents the effect of the number of fixed objects on the success rates of the planners. For a constant number of objects in the scene if we increase the number of fixed objects, the performance of our algorithm increases owing to the reduction in the dimensionality of the problem. However, increasing the fixed objects also invalidates a lot more actions. This implies that in the non-selective case more actions have to be simulated to find a valid path.

The  utility of the algorithm was also demonstrated through real world experiments on the 7 DOF manipulator of UBTech’s Walker robot. A video of the real-world experiments and the simulation results can be found \href{https://youtu.be/OmUEgUzGizY}{here}.

\section{Conclusion}
In this paper we present a technique for intelligently integrating physics based simulators into search-based planners for the domain of manipulation among movable objects through non-prehensile maneuvers. Our approach selectively simulates actions by identifying relevant objects in the environment. This significantly speeds up planning without sacrificing completeness and suboptimality guarantees. We plan on extending this idea to domains that go beyond pick-and-place tasks, like utilizing tools. Further, we would like to address the occasional inaccuracies in simulation modelling by learning from real robot experiences.

\bibliographystyle{IEEEtran}
\bibliography{IEEEabrv}

\end{document}